\documentclass[11pt,a4paper]{article}
\usepackage[hyperref]{acl2021}
\usepackage{times}
\usepackage{latexsym}
\usepackage{enumitem}

\usepackage{lipsum,multicol}
\usepackage{microtype}
\usepackage{amsmath,array,graphicx}
\usepackage{float} 
\usepackage{multirow}

\aclfinalcopy 
\def\aclpaperid{4227} 

\title{N-Best ASR Transformer: Enhancing SLU Performance using Multiple ASR Hypotheses}

\author{
  Karthik Ganesan\thanks{\hspace{1.5 mm} The first three authors have equal contribution.}\hspace{1.5 mm}, Pakhi Bamdev\footnotemark[1]\hspace{1.5 mm}, Jaivarsan B\footnotemark[1]\hspace{1.5 mm}, Amresh Venugopal, Abhinav Tushar \\
  Vernacular.ai\\
  \texttt{\{karthikganesan17, pakhi.bamdev.in\}@gmail.com}
  \\\texttt{\{jaivarsan, amresh, abhinav\}@vernacular.ai}
}
\date{}

\begin{document}
\maketitle
\begin{abstract}
Spoken Language Understanding (SLU) systems parse speech into semantic structures like dialog acts and slots. This involves the use of an Automatic Speech Recognizer (ASR) to transcribe speech into multiple text alternatives (hypotheses). Transcription errors, common in ASRs, impact downstream SLU performance negatively. Approaches to mitigate such errors involve using richer information from the ASR, either in form of N-best hypotheses or word-lattices. We hypothesize that transformer models learn better with a simpler utterance representation using the concatenation of the N-best ASR alternatives, where each alternative is separated by a special delimiter [SEP]. In our work, we test our hypothesis by using concatenated N-best ASR alternatives as the input to transformer encoder models, namely BERT and XLM-RoBERTa, and achieve performance equivalent to the prior state-of-the-art model on DSTC2 dataset.  We also show that our approach significantly outperforms the prior state-of-the-art when subjected to the low data regime. Additionally, this methodology is accessible to users of third-party ASR APIs which do not provide word-lattice information.
\end{abstract}

\section{Introduction}
Spoken Language Understanding (SLU) systems are an integral part of Spoken Dialog Systems. They parse spoken utterances into corresponding semantic structures e.g. dialog acts. For this, a spoken utterance is usually first transcribed into text via an Automated Speech Recognition (ASR) module. Often these ASR transcriptions are noisy and erroneous. This can heavily impact the performance of downstream tasks performed by the SLU systems.

To counter the effects of ASR errors, SLU systems can utilise additional feature inputs from ASR. A common approach is to use N-best hypotheses where multiple ranked ASR hypotheses are used, instead of only 1 ASR hypothesis. A few ASR systems also provide additional information like word-lattices and word confusion networks. Word-lattice information represents alternative word-sequences that are likely for a particular utterance, while word confusion networks are an alternative topology for representing a lattice where the lattice has been transformed into a linear graph. Additionally, dialog context can help in resolving ambiguities in parses and reducing impact of ASR noise. 

\textbf{N-best hypotheses: } \citet{li2019robust} work with 1-best ASR hypothesis and exploits unsupervised ASR error adaption method to map ASR hypotheses and  transcripts to a similar feature space. On the other hand, \citet{khan2015hypotheses} uses multiple ASR hypotheses to predict multiple semantic frames per ASR choice and determine the true spoken dialog system's output using additional context. \textbf{Word-lattices: }\citet{ladhak2016latticernn} propose using recurrent neural networks (RNNs) to process weighted lattices as input to SLU. \citet{vsvec2015word} presents a method for converting word-based ASR lattices into word-semantic (W-SE) which reduces the sparsity of the training data. \citet{huang2019adapting} provides an approach for adapting lattices with pre-trained transformers. \textbf{Word confusion networks (WCN): } \citet{jagfeld-vu-2017-encoding} proposes a technique to exploit word confusion networks (WCNs) as training or testing units for slot filling.  \citet{masumura2018neural} models WCN as sequence of bag-of-weighted-arcs and introduce a mechanism that converts the bag-of-weighted-arcs into a continuous representation to build a neural network based spoken utterance classification. \citet{DBLP:conf/interspeech/LiuZZC0020} proposes a BERT based SLU model to encode WCNs and the dialog context jointly to reduce ambiguity from ASR errors and improve SLU performance with pre-trained models.

The motivation of this paper is to improve performance on downstream SLU tasks by exploiting \textit{transfer learning} capabilities of the pre-trained transformer models. Richer information representations like word-lattices (\citet{huang2019adapting}) and word confusion networks (\citet{DBLP:conf/interspeech/LiuZZC0020}) have been used with GPT and BERT respectively. These representations are non-native to Transformer models, that are pre-trained on plain text sequences. We hypothesize that transformer models will learn better with a simpler utterance representation using concatenation of the N-best ASR hypotheses, where each hypothesis is separated by a special delimiter [SEP]. We test the effectiveness of our approach on a dialog state tracking dataset - DSTC2 \cite{henderson2014second}, which is a standard benchmark for SLU.

\textbf{Contributions: }(i) Our proposed approach, trained with a simple input representation, exceeds the competitive baselines in terms of accuracy and shows equivalent performance on the F1-score to the prior state-of-the-art model. (ii) We significantly outperform the prior state-of-the-art model in the low data regime. We attribute this to the effective \textit{transfer learning} from the pre-trained Transformer model. (iii) This approach is accessible to users of third party ASR APIs unlike the methods that use word-lattices and word confusion networks which need deeper access to the ASR system.

\section{N-Best ASR Transformer}
\textit{N-Best ASR Transformer}\textsuperscript{1}\footnotetext[1]{ The code is available at \href{https://github.com/Vernacular-ai/N-Best-ASR-Transformer}{https://github.com/Vernacular-ai/N-Best-ASR-Transformer}} works with a simple input representation achieved by concatenating the N-Best ASR hypotheses together with the dialog context (system utterance). Pre-trained transformer models, specifically BERT and XLMRoBERTa, are used to encode the input representation. For output layer, we use a semantic tuple classifier (STC) to predict \textit{act-slot-value} triplets. The following sub-sections describe our approach in detail.

\subsection{Input Representation}
\label{sec:input}
For representing the input we concatenate the last system utterance $S$ (dialog context), and the user utterance $U$. $U$ is represented as concatenation of the N-best\textsuperscript{2}\footnotetext[2]
{We use ASR transcriptions (N $\leq$ 10)  provided by DSTC2 dataset to perform our experiments. Our input structure can support variable N during training and inference.} ASR hypotheses, separated by a special delimiter, [SEP]. The final representation is shown in equation \ref{eq:sys_utt} below:
\begin{equation}
x_{i} = [\text{CLS}] \oplus \text{TOK}(S_{i}) \oplus \bigoplus_{j=1}^{N} (\text{TOK}(U_{i}^{j}) \oplus [\text{SEP}])
\label{eq:sys_utt}
\end{equation}

Here, $U_{i}^{j}$ refers to the $j^{th}$ ASR hypothesis for the $i^{th}$ sample, $\oplus$ denotes the concatenation operator, TOK(.) is the tokenizer, [CLS] and [SEP] are the special tokens. 

\begin{figure}[H]
    \centering
    \includegraphics[width=7.7cm, height = 0.75cm]{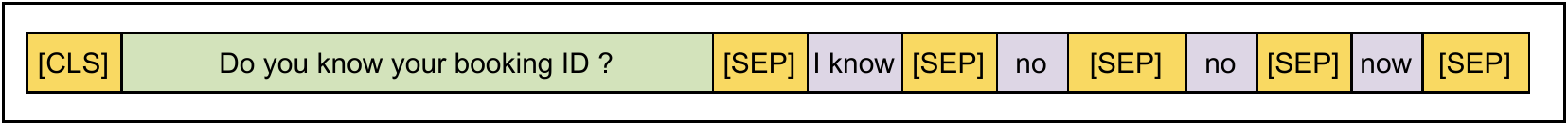}
    \caption{Input representation: The green boxes represents the last system utterances followed by ASR hypotheses of user utterances concatenated together with a [SEP] token.}
    \label{fig:sys_utt}
\end{figure}

\begin{figure*}
    \includegraphics[width=16cm, height = 7cm]{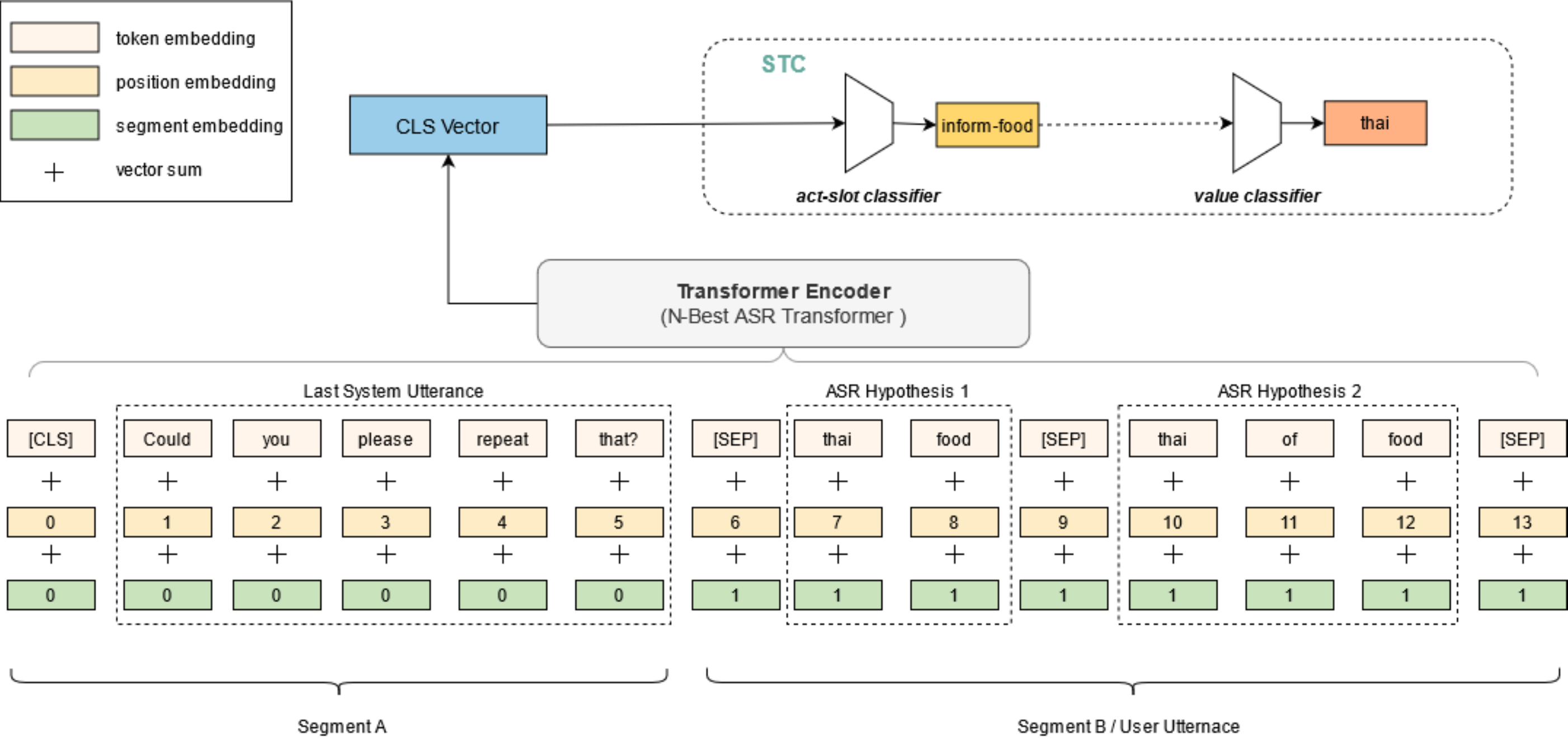}
    \caption{\textbf{N-Best ASR Transformer:} The input representation is encoded by a transformer model which forms an input for a Semantic Tuple Classifier (STC). STC uses binary classifiers to predict the presence of act-slot pairs, followed by a multi-class classifier that predicts the value for each act-slot pair. 
}
    \label{fig:arch}
    \vskip -3ex
\end{figure*}

As represented in figure \ref{fig:arch}, we also pass segment IDs along with the input to differentiate between segment $a$ (last system utterance) and segment $b$ (user utterance).

\subsection{Transformer Encoder}
The above mentioned input representation can be easily used with any pre-trained transformer model. For our experiments, we select BERT \cite{devlin-etal-2019-bert} and XLM-RoBERTa\textsuperscript{3}\footnotetext[3]
{The model name XLM-RoBERTa and XLM-R will be used interchangeably throughout the paper.} \cite{conneau-etal-2020-unsupervised} for their recent popularity in NLP research community. 

\subsection{Output Representation}
The final hidden state of the transformer encoder corresponding to the special classification token [CLS] is used as an aggregated input representation for the downstream classification task by a semantic tuple classifier (STC) \citep{mairesse2009spoken}. STC uses two classifiers to predict the \textit{act-slot-value} for a user utterance. A binary classifier is used to predict the presence of each \textit{act-slot} pair, and a multi-class classifier is used to predict the \textit{value}  corresponding to the predicted act-slot pairs. We omit the latter classifier for the act-slot pairs with no value (like \emph{goodbye}, \emph{thankyou}, \emph{request\_food} etc.).

\section{Experimental Setup}
\subsection{Dataset}
We perform our experiments on data released by the Dialog State Tracking Challenge (DSTC2) \cite{henderson2014second}. It includes pairs of utterances and the corresponding set of \textit{act-slot-value} triplets for training (11,677 samples), development (3,934 samples), and testing (9,890 samples). The task in the dataset is to parse the user utterances like ``\emph{I  want a moderately priced restaurant.}" into a corresponding semantic representation in the form of ``\textit{inform(pricerange=moderate)}" triplet. For each utterance, both the manual transcription and a maximum of 10-best ASR hypotheses are provided. The utterances are annotated with multiple \textit{act-slot-value} triplets. For transcribing the utterances DSTC2 uses two ASRs - one with an artificially degraded statistical acoustic model, and one which is fully optimized for the domain. Training and development sets include transcriptions from both the ASRs.  To utilise this dataset we first transform it into the input format as discussed in section \ref{sec:input}.

\subsection{Baselines}
\label{sec:baseline}
We compare our approach with the following baselines:
\begin{itemize}
    \item \textbf{SLU2} \cite{williams2014web}: Two binary classifiers (decision trees) are used with word n-grams from the ASR N-best list and the word confusion network. One predicts the presence of that slot-value pair in the utterance and the other estimate for each user dialog act.
    
    \item \textbf{CNN+LSTMw4}  \cite{rojas-barahona-etal-2016-exploiting}: A convolution neural network (CNN) is trained with the N-best ASR hypotheses to output the utterance representation. A long-short term memory network (LSTM) with a context window size of 4 outputs a context representation.  The models are jointly trained to predict for the act-slot pair. Another model with the same architecture is trained to predict for the value corresponding to the predicted act-slot pair. 

    \item \textbf{CNN} \cite{zhao2018improving}: Proposes CNN based models for dialog act and slot-type prediction using 1-best ASR hypothesis.
    
    \item \textbf{Hierarchical Decoding} \cite{zhao2019hierarchical}: A neural-network based binary classifier is used to predict the act and slot type. A hybrid of sequence-to-sequence model with attention and pointer network is used to predict the value corresponding to the detected act-slot pair.1-Best ASR hypothesis was used for both training and evaluation tasks.
    
    \item \textbf{WCN-BERT + STC} \cite{DBLP:conf/interspeech/LiuZZC0020}:  Input utterance is encoded using the Word Confusion Network (WCN) using BERT by having the same position ids for all words in the bin of a lattice and modifying self-attention to work with word probabilities. A semantic tuple classifier uses a binary classifier to predict the act-slot value, followed by a multi-class classifier that predicts the value corresponding to the act-slot tuple.

\end{itemize}

\subsection{Experimental Settings}
We perform hyper-parameter tuning on the validation set to get optimal values for dropout rate $\delta$, learning rate $lr$, and the batch size $b$. Based on the best F1-score, the final selected parameters were $\delta$ = 0.3,  $lr$ = 3e-5 and $b$  = 16. We set the warm-up rate $wr$ = 0.1, and L2 weight decay  $L2$ = 0.01. We make use of Huggingface's \emph{Transformers} library \cite{wolf-etal-2020-transformers} to fine-tune the \textit{bert-base-uncased} and \textit{xlm-roberta-base}, which is optimized over Huggingface’s BertAdam optimizer. We trained the model on Nvidia T4 single GPU on AWS EC2 g4dn.2xlarge instance for 50 epochs. We apply early stopping and save the best-performing model based on its performance on the validation set. 

\section{Results}
In this section, we compare the performance of our approach with the baselines on the DSTC2 dataset. To compare the \textit{transfer learning} effectiveness of pre-trained transformers with \textit{N-Best ASR BERT} (our approach) and the previous state-of-the-art model \textit{WCN-BERT STC}, we perform comparative analysis in the low data regime. Additionally, we perform an ablation study on \textit{N-Best ASR BERT} to see the impact of modeling dialog context (last system utterance) with the user utterances.

\subsection{Performance Evaluation}
\begin{table}[H]
\centering
\resizebox{\columnwidth}{!}{%
\begin{tabular}{lcc}
\hline
\textbf{Model}             & \textbf{F1-score} & \textbf{Accuracy} \\ \hline
SLU2                       & 82.1              & -                 \\
CNN+LSTM\_w4               & 83.6              & -                 \\
CNN                        & 85.3              & -                 \\
Hierarchical Decoding      & 86.9              & -                 \\
WCN-BERT + STC             & \textbf{87.9}     & 81.1              \\ \hline
\textbf{N-Best ASR XLM-R (Ours)} & 87.4     & \textbf{81.9}     \\
\textbf{N-Best ASR BERT (Ours) } & 87.8     & 81.8              \\ \hline
\end{tabular}}
\caption{F1-scores (\%) and utterance-level accuracy (\%) of
baseline models and our proposed model on the test set.}
\label{table:main_results}
\end{table}

Since the task is a multi-label classification of \textit{act-slot-value} triplets, we report utterance level accuracy and F1-score. A prediction is correct if the set of labels predicted for a sample exactly matches the corresponding set of labels in the ground truth. As shown in Table \ref{table:main_results}, we compare our models, \textit{N-Best ASR BERT} and \textit{N-Best ASR XLM-R}, with baselines mentioned in section \href{sec:baseline}. Both of our proposed models, trained with concatenated N-Best ASR hypotheses, outperform the competitive baselines in terms of accuracy and show comparable performance on F1-score with \textit{WCN-BERT STC}.

\subsection{Performance in Low Data Regime}
\begin{table}[H]
\centering
\resizebox{\columnwidth}{!}{%
\begin{tabular}{lcc}
\hline
\textbf{Train Data (\%age)} & \textbf{WCN-BERT STC} & \textbf{\textit{N-Best ASR BERT}} \\ \hline
5                 & 78.5              & \textbf{83.9}                       \\
10                & 80.3              & \textbf{85.5}                       \\
20                & 84.4              & \textbf{86.7}                       \\
50                & 85.9              & \textbf{87.7}                       \\ \hline
\end{tabular}}
\caption{F1-scores (\%) for our proposed model \textit{N-Best ASR BERT} (ours) and \textit{WCN-BERT STC} (previous state-of-the-art.}
\label{table:samp_complex}
\end{table}
To study the performance of model in the low data regime, we randomly select $p$ percentage of samples from the training set in a stratified fashion, where $p$ $\in$ \{5, 10, 20, 50\}. We pick our model \textit{N-Best ASR BERT} and  \textit{WCN-BERT STC} for this study because both use BERT as the encoder model. For both models, we perform experiments using the same training, development, and testing splits. From Table \ref{table:samp_complex}, we find that \textit{N-Best ASR BERT} outperforms \textit{WCN-BERT STC} model significantly for low data regime, especially when trained on 5\% and 10\% of the training data. It shows that our approach effectively \textit{transfer learns} from pre-trained transformer's knowledge. We believe this is due to the structural similarity between our input representation and the input BERT was pre-trained on.

\subsection{Significance of Dialog Context}
\begin{table}[H]
\centering
\resizebox{\columnwidth}{!}{%
\begin{tabular}{llcc}
\hline
\textbf{Model}                       & \textbf{Variation}       & \multicolumn{1}{l}{\textbf{F1-score}} & \multicolumn{1}{l}{\textbf{Accuracy}} \\ \hline
\multirow{2}{*}{N-Best ASR BERT} & without system utterance & 86.5                                & 80.2                                 \\
                                    & with system utterance    & \textbf{87.8}                        & \textbf{81.8}                        \\ \hline
\end{tabular}}
\caption{F1-scores (\%) and utterance-level accuracy (\%) of our model \textit{N-Best ASR BERT} on the test set when trained with and without system utterances.}
\label{table:sys_utt_ablation}
\end{table}

Through this ablation study, we try to understand the impact of dialog context on model's performance. For this, we train \textit{N-Best ASR BERT} in the following two settings:
\begin{itemize}[itemsep=-4pt]
\item When input representation consists of only the user utterance.
\item When input representation consists of both the last system utterance (dialog context) and the user utterance as shown in figure \ref{fig:sys_utt}. 
\end{itemize}

As presented in Table \ref{table:sys_utt_ablation}, we observe that modeling the last system utterance helps in achieving better F1  and utterance-level accuracy by the difference of 1.3\% and 1.6\% respectively. 

\begin{figure}[H]
    \centering
    \includegraphics[width=7.5cm, height = 1.6cm]{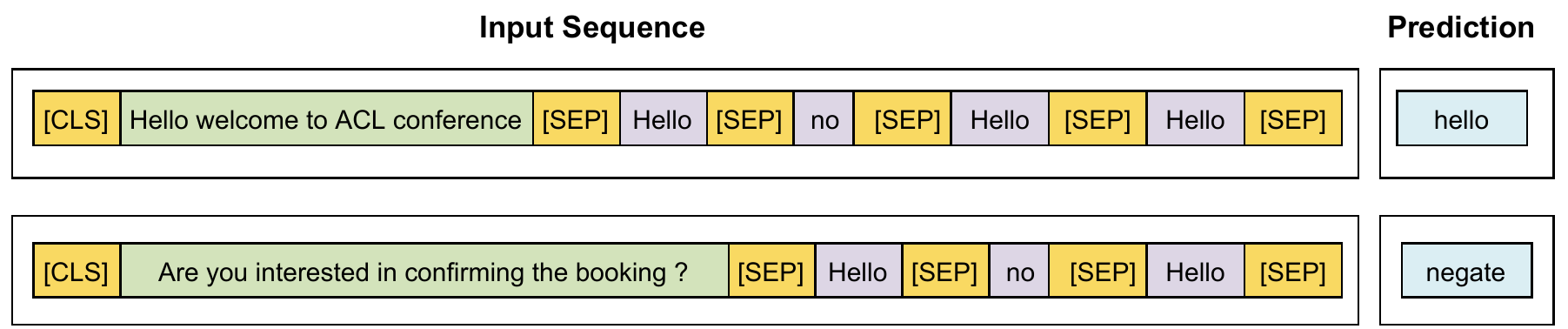}
    \caption{Significance of Dialog Context: The green box depicts the dialog context that helps disambiguate the very similar ASR hypotheses shown in purple boxes. }
    \label{fig:sys_utt}
\end{figure}
 It proves that dialog context helps in improving the performance of downstream SLU tasks. Figure \ref{fig:sys_utt} represents one such example where having dialog context in form of the last system utterance helps disambiguate between the two similar user utterances.

\section{Conclusion}
In this work, building on a simple input representation, we propose \textit{N-Best ASR Transformer}, which outperforms all the competitive baselines on utterance-level accuracy for the DSTC2 dataset. However, the highlight of our work is in achieving significantly higher performance in an extremely low data regime. This approach is accessible to users of third-party ASR APIs, unlike the methods that use word-lattices and word confusion networks. 
As future extensions to this work,  we plan to :
\begin{itemize}[itemsep=-2pt]
    \item Enable our proposed model to generalize to out-of-vocabulary (OOV) slot values.
    \item Evaluate our approach in a multi-lingual setting.
     \item Evaluate on different values N in N-best ASR.
    \item Compare the performance of our approach on ASRs with different Word Error Rates (WERs).
\end{itemize} 

\section*{Acknowledgement}
We are highly grateful to our organization \href{https://vernacular.ai/}{Vernacular.ai} and our Machine Learning Team for (i) exposing us to practical problems related to multilingual voice-bots, (ii) giving us access to resources to solve this problem, (iii) helping us deploy this work in production for real-world users, and (iv) for their excellent feedback on this work.

\bibliographystyle{acl_natbib}
\bibliography{anthology,acl2021}


\end{document}